\documentclass[sigconf]{acmart}
\AtBeginDocument{%
  \providecommand\BibTeX{{%
    \normalfont B\kern-0.5em{\scshape i\kern-0.25em b}\kern-0.8em\TeX}}}

\acmConference[ICSE 2022]{The 44th International Conference on Software Engineering}{May 21–29, 2022}{Pittsburgh, PA, USA}

\copyrightyear{2022}
\acmYear{2022}
\setcopyright{rightsretained}
\acmConference[ICSE '22]{44th International Conference on Software
Engineering}{May 21--29, 2022}{Pittsburgh, PA, USA}
\acmBooktitle{44th International Conference on Software Engineering
(ICSE '22), May 21--29, 2022, Pittsburgh, PA, USA}
\acmDOI{10.1145/3510003.3510095}
\acmISBN{978-1-4503-9221-1/22/05}

\usepackage{xcolor}
\usepackage[ruled]{algorithm2e}
\usepackage{amsmath,amsthm}
\usepackage{listings}
\usepackage{multicol}
\usepackage[font={small,it}]{caption}
\usepackage{fancybox}

\begin{document}

\setlength{\pdfpageheight}{\paperheight}
\setlength{\pdfpagewidth}{\paperwidth}





\title{DeepStability: A Study of Unstable Numerical Methods and Their Solutions in Deep Learning}

\author{Eliska Kloberdanz}
\affiliation{Department of Computer Science\\Iowa State University}
\email{eklober@iastate.edu}

\author{Kyle G. Kloberdanz}
\affiliation{Cape Privacy}
\email{kyle.g.kloberdanz@gmail.com}

\author{Wei Le}
\affiliation{Department of Computer Science\\Iowa State University}
\email{weile@iastate.edu}

\begin{abstract}

Deep learning (DL) has become an integral part of solutions to various important problems, which is why ensuring the quality of DL systems is essential. One of the challenges of achieving reliability and robustness of DL software is to ensure that algorithm implementations are {\it numerically stable}. DL algorithms require a large amount and a wide variety of numerical computations. A naive implementation of numerical computation can lead to errors that may result in incorrect or inaccurate learning and results. A numerical algorithm or a mathematical formula can have several implementations that are mathematically equivalent, but have different numerical stability properties. Designing numerically stable algorithm implementations is challenging, because it requires an interdisciplinary knowledge of software engineering, DL, and numerical analysis. In this paper, we study two mature DL libraries PyTorch and Tensorflow with the goal of identifying unstable numerical methods and their solutions. Specifically, we investigate which DL algorithms are numerically unstable and conduct an in-depth analysis of the root cause, manifestation, and patches to numerical instabilities. Based on these findings, we launch {\it DeepStability}, the first database of numerical stability issues and solutions in DL. Our findings and {\it DeepStability} provide future references to developers and tool builders to prevent, detect, localize and fix numerically unstable algorithm implementations. To demonstrate that, using {\it DeepStability} we have located numerical stability issues in Tensorflow, and submitted a fix which has been accepted and merged in.

\end{abstract}



\maketitle

\section{Introduction}
Deep learning (DL) has become an integral part of solutions to various important problems such as navigation of driverless cars, natural language processing for language translation, credit card fraud detection, or automated trading. Ensuring the quality of a deep learning system is an essential task. One of the challenges of achieving reliability and robustness of DL is ensuring that algorithms implementations are {\it numerically stable}. In traditional numerical analysis literature, numerical stability is treated as a property of algorithms. An unstable numerical method produces large changes in outputs for small changes in inputs ~\cite{Jong1977TowardsAF}, which can lead to unexpected outputs or errors. Especially in the implementation of DL, we rely on high precision floating point computations to reach reliable decisions and need to use large integers to process very large datasets, which are ubiquitous in practice. As a result, unstable methods can trigger overflow or underflow and truncation. Such errors are then propagated through iterations of training, leading to low quality models and wasting computational resources. For example, when DL is deployed in autonomous vehicles, incorrect numerical computations can lead to incorrect vehicle trajectory, turning, lane positioning and navigation, resulting in severe consequences \cite{wakabayashi_2018, berboucha_2018, vlasic_boudette_2016}. 


Implementing numerically stable algorithms is challenging. A numerical algorithm or a mathematical formula can have several implementations that are mathematically equivalent, but have very different numerical stability properties. To design a stable implementation, DL developers need to have an in-depth interdisciplinary knowledge of mathematics, DL algorithms, numerical analysis, computer programming, and finite precision floating point computation \cite{Franco2017ACS}. There are some general guidelines that can be followed to develop numerically stable algorithms; however, for each numerical method, we need to have specific solutions to mitigate numerical instability \cite{Higham2002AccuracyAS}. In addition, detecting and diagnosing numerical stability issues is hard for two reasons. First, similar to security vulnerabilities, numerical stability issues can be triggered only by a special small range of inputs; however, such issues can be consequential, e.g., failed training or incorrect predictions in a safety-critical DL systems. Second, numerical stability errors sometimes occur silently and are not observed until they propagate through iterations of training far from the source of instability. 


Prior software engineering solutions to numerical instability, including detection  \cite{Lee2015RAIVERA, Fu2015AutomatedBE}, automated repair \cite{Yi2019EfficientAR}, debugging \cite{Chowdhary2020DebuggingAD}, and increased precision computations \cite{Benz2012ADP}, mostly focus on code but not algorithms. For example, \cite{Bao2013OntheflyDO} monitor whether relative error becomes inflated during program execution to detect code segments with risks of numerical instability, and then automatically switch to high precision computations. They improve upon \cite{Benz2012ADP} in terms of speed; however, increasing precision is not the only or always appropriate solution.  Mathematical solutions that involve crafting more numerically stable algorithm implementations can provide more reliable and speed efficient solutions to numerical instability. 


Considering increasing applications of DL in industry, more and more developers will need to implement DL numerical algorithms.  More tools should be developed to help detect, test, diagnose and repair numerical instability.  Numerical stability issues are different from any bugs in traditional software. Also, the code patterns, patches and root causes of numerical instability are very specific to algorithms, and to the best of our knowledge, such information of DL stability has not been covered in the numerical analysis literature. Thus, the goal of this paper is to discover the state-of-the-art knowledge of numerical stability, such as unstable methods and their solutions, in the domain of DL to support DL developers to write more numerically stable code. Our work studied numerical stability from the DL algorithmic, mathematical and code perspectives, which has not been done in the previous research.




Specifically, in this paper, we discover and analyze a comprehensive list of unstable numerical methods used in DL algorithms and prepare a repository of their solutions. We studied two mature and important DL libraries, PyTorch and Tensorflow. Through analyzing their commit histories, we distilled the patches, unit tests and new features related to unstable methods and their solutions. We have cataloged the data collected, a total of 252 entries, in a database, called {\it DeepStability}. It is publicly available~\footnote{https://deepstability.github.io/} and will serve as a starting point, where we can continuously add unstable methods and their solutions for references and reuses. Fixing unstable implementations can be hard and time-consuming. {\it DeepStability} can educate developers which algorithms/math formulas have numerical issues and avoid introducing unstable implementations. {\it DeepStability} can also help diagnose and fix numerical stability bugs, e.g., a developer can cite {\it DeepStability} in their pull request for numerical stability fixes. This way they can both show why the previous version of an algorithm has numerical stability bugs and the example code for a fix. Using {\it DeepStability}, we have located a numerical stability issue in Tensorflow, and submitted a fix~\footnote{https://github.com/tensorflow/tensorflow/pull/50855} which has been accepted and merged in.

We found that numerical stability issues indeed widely existed and were discussed through the DL development process. Using the data we collected, we investigated what DL algorithms are susceptible to numerical instability (RQ1), what is the root cause and impact of unstable methods (RQ2), and what solutions fix numerical instabilities (RQ3). From the real-world data we analyzed, we discovered new numerical vulnerability patterns that, to the best of our knowledge, have not been reported in the literature.




In summary, the main research contributions of this paper are:
\begin{enumerate}
 
 \item We classified which DL algorithms are susceptible to numerical instability and explained why;  (\S 4.1) 
 
 \item We performed an in-depth analysis of the root cause and impact of numerical stability bugs in DL algorithms; (\S 4.2)
 
 \item We summarized both mathematical and code level solutions for numerically stable DL algorithms;  (\S 4.3)
 
 \item We discovered new unstable methods and their solutions in deep learning that are not discussed in prior literature; (\S 5)
 
\item We launched DeepStability, the first database of numerical stability issues and solutions in DL as a reference for DL developers and tool builders  (\url{https://deepstability.github.io}).


\end{enumerate}

\section{A Motivating Example}

In this section, we show an unstable numerical method {\it softmax} and explain why mathematically equivalent operations can have different numerical stability properties that can lead to undesirable outcomes. {\it softmax} is commonly used in various multi-class classification algorithms such as logistic regression, linear discriminant analysis, naive Bayes classifier, and artificial neural networks, including DNN, CNN, RNN and GAN. Additionally, {\it softmax} is also used in reinforcement learning for converting value functions into action probabilities. This motivating example demonstrates the importance of understanding numerical stability, which is required for correct implementation of DL algorithms.


\subsection{Numerically Unstable Softmax}
Softmax is a normalized exponential function that takes a vector of $n$ real values as input and outputs a vector of $n$ real values that represent a probability distribution and sum up to 1. In DL classifiers, softmax is used in the last neural network layer, because it normalizes the output of the prior network layer, a vector of size $n$, to a probability distribution over $n$ predicted output classes. 

\begin{equation}
\label{softmax_unstable}
softmax(\vec{x})_i = \frac{e^{x_i}}{\sum_{j=1}^{n} e^{x_j}} 
\end{equation}

The definition of \textit{softmax} given in Equation \ref{softmax_unstable} and its C++ implementation at lines~1--12 in Listing 1 are numerically unstable. When given an input vector x=[10.0, 100.0, 1000.0], $e^{100.0}$ and $e^{1000.0}$ overflow and are set to an $inf$ at lines~6 and 9 . Hence, {\tt sum} is computed as $22026.5 + inf + inf = inf$ at line~6. As a consequence, {\tt result[j]} returns $\frac{inf}{inf} = -nan$ at line~9. Similarly, when given an input vector y=[-1000.0, -10000.0, -1000000.0], $e^{-1000.0}$, $e^{-10000.0}$, and $e^{-1000000.0}$ underflow and are set to zero at lines~6 and 9. Therefore, {\tt sum} is computed as $0 + 0 + 0 = 0$ at line~6. This results in a divide by zero on line~9, which is an invalid operation that yields a \textit{NaN}. In both cases \textit{softmax} becomes undefined and will cease to output meaningful probabilities. 



\begin{lstlisting}[basicstyle = \small, frameround=\thinlines, caption = Unstable and Stable Implementations of Softmax, language=C++, numbers=left, fontadjust]
vector<float> softmax_unstable(const vector<float> &x) 
{
    float sum = 0;
    vector<float> result;
    result.resize(x.size());
    for (size_t i = 0; i < x.size(); i++) {
        sum += exp(x[i]);
    }
    for (size_t j = 0; j < x.size(); j++) {
        result[j] = exp(x[j]) / sum;
    }
    return result;
}
vector<float> softmax_stable(const vector<float> &x) 
{
    float sum = 0;
    vector<float> result;
    result.resize(x.size());
    float max = *max_element(x.begin(), x.end());
    for (size_t i = 0; i < x.size(); i++) {
        sum += exp(x[i] - max);
    }
    for (size_t j = 0; j < x.size(); j++) {
        result[j] = exp(x[j] - max) / sum;
    }
    return result;
}
Unstable softmax of x=[10.0,100.0,1000.0]: 
0, -nan, -nan
Stable softmax of x=[10.0,100.0,1000.0]: 
0, 0, 1

Unstable softmax of y=[-1000.0,-10000.0,-1000000.0]: 
-nan, -nan, -nan
Stable softmax of y=[-1000.0,-10000.0,-1000000.0]: 
1, 0, 0
\end{lstlisting}


    



\subsection{Error Propagation in DL Algorithms}
The overflow and underflow in \textit{softmax} caused by numerical instability propagates through neural network and causes it to stop learning. We performed experiments with the unstable \textit{softmax} implementation from Listing 1 on MNIST to demonstrate that. Given a fully connected deep neural network (DNN) with standard learning parameters but numerically unstable \textit{softmax} function in its last layer, we observe that after a couple of training epochs weights, biases, and loss become \textit{NaN}. The source of this issue are rounding errors caused by numerical instability in class probabilities computed by \textit{softmax} forward pass. Given ten different classes, we observe the following probabilities shown in Listing 2.

\begin{lstlisting}[basicstyle = \small, frameround=\thinlines, caption = Underflow in Class Probabilities due to Numerically Unstable Softmax, fontadjust]

Class  1: 1.2295200002966873e-129
Class  2: 0.0 //underflow
Class  3: 1.3695324610698374e-266
Class  4: 5.966951373841794e-250
Class  5: 2.5766327266617867e-89
Class  6: 3.4522175477266625e-234
Class  7: 1.3344020481166367e-162
Class  8: 1.7656178002771016e-269
Class  9: 1.0
Class 10: 1.616e-321
\end{lstlisting}

Listing 2 indicates that the DNN is very confident that the input example is of class 9. In fact, it reports that the probability of class 9 equals 1.0, while the probabilities of other classes are extremely small. Due to rounding error, the probability of class 2 underflows and becomes 0.0. The operation that follows softmax forward pass is softmax backward pass, which calculates the derivative of loss w.r.t. the softmax output as follows: $- y\_true/softmax\_output$. In this formula $y\_true$ is hot-encoded correct labels, i.e.: [0 0 0 0 0 0 0 0 1 0], which represent the 10 possible labels and identify class 9 as the correct one. Since $softmax\_output$ for class 2 is zero, this will cause a divide-by-zero in $- y\_true/softmax\_output$, an invalid operation that outputs a \textit{NaN}. As a result, the gradient vector of softmax will contain a \textit{NaN} which will propagate through the network and cause \textit{NaNs} in weights and biases. \textit{NaNs} in weights and biases will in turn cause the output of the next forward pass to become a \textit{NaN}, which will then cause the loss to became a NaN. 

Therefore, a single error originating from a numerically unstable implementation of the softmax function can create a snowball effect and prevent the network from learning. Unfortunately, deep learning APIs such as Keras or PyTorch continue training even when the network parameters become NaN, which is a waste of computational resources and the developer's time.

\subsection{Numerically Stable Solution}
To mitigate numerical stability issues of \textit{softmax} discussed above, we can rewrite the unstable formula in Equation \ref{softmax_unstable} to its mathematically equivalent, but more numerically stable version shown in Equation \ref{stablesoftmax} and implemented in C++ on lines 13-24 in Listing 1.
\vspace{-2pt}
\begin{equation}
\label{stablesoftmax}
    softmax(\vec{x})_i = \frac{e^{x_i - max(\vec{x})}}{\sum_{j=1}^{n} e^{x_j-max(\vec{x})}}
\end{equation}
\vspace{-2pt}
This solution normalizes inputs to ensure that they are not too large or too small; and therefore, decreases the risk of arithmetic exceptions. Specifically, $max(\vec{x})$  returns the largest scalar element in vector $\vec{x}$. Subtracting $max(\vec{x})$ from each $x_i$ element of vector $\vec{x}$ implies two properties. First, the largest input $x_i$ is passed into the exponential function as a zero. Second, at least one value in the summation in the denominator is equal to 1, because the largest $x_i$ is passed in as $x_i = 0$ and $exp(0) = 1$. The first property decreases the risk of overflow and the second one prevents underflow in the denominator that would result in division by zero, an invalid operation.  
Lines 27 and 30 show that the this stable version of softmax yields correct outputs. In our appendix, we show that the two versions of softmax in Equations \ref{softmax_unstable} and \ref{stablesoftmax} are mathematically equivalent. Additionally, the appendix also contains a similar proof for logsoftmax. Prior literature shows how to rewrite logsoftmax to obtain a numerically stable solution, but does not provide a step by step proof that the two formulas are mathematically equivalent.




\section{Study Methods}
\subsection{Study Goal and Process}
To identify and analyze unstable numerical methods of DL algorithms similar to {\it softmax} presented in Section 2, we studied PyTorch and Tensorflow code repositories. Our goal is to inspect commits that are related to numerical stability to localize patches, tests, and any other code additions related to the important numerical methods in DL. We selected PyTorch and Tensorflow for our study, because they are the most widely used, well-maintained and mature DL libraries~\cite{Kumar2020AdversarialML}. Thus, we believe that the problems and solutions we discovered here are representative and can be reused in other DL implementations.

Our approach for selecting keywords to search PyTorch and Tensorflow focuses on finding numerical stability commits while avoiding excessive noise (irrelevant commits). We used the keywords from \cite{Franco2017ACS} as a reference and prepared a list of keywords that can indicate the symptoms of numerical stability such as “overflow, underflow, precision, NaN, inf”, keywords that indicate when numerical stability can occur e.g., “numerical”, “approximation”, “zero”, and keywords that may describe stability issues such as “stable, stability”. Finally, we used sample search results to further refine the keywords that can return relevant commits and minimize noise.

Using the following keywords we found 189 commits in PyTorch: "stability", "stable", "numerical", "approximation", "overflow", "underflow", "precision", "NaN", "zero", and "inf". These commits were manually analyzed to asses whether they relate to numerical stability, which reduced the number of relevant commits to 123. The same keywords yielded 696 commits in Tensorflow, a significantly higher number than in PyTorch. We observed that these keywords brought in many irrelevant commits that polluted the search results. For example, "stability" search results typically related to stable software releases, not numerical stability. Thus, we further refined the keywords for searching Tensorflow. The following keywords yielded 307 commits: "numerical stability", "numerically stable", "stable", "unstable", "overflow", "underflow". We analyzed these commits and filtered out the ones that did not relate to numerical stability, which yielded 129 commits. Therefore, the total number of commits in PyTorch and Tesorflow relevant to numerical stability came to 252.


 When inspecting the commits, we follow a methodology used in prior software engineering works that studied numerical bugs \cite{Franco2017ACS}. The two authors conducted an independent analysis and then discussed their results. If an agreement could not be reached or both authors were not confident in their analysis, commits were excluded. For 6/258 (2.3\%) of commits, the author(s) had low confidence on what was the numerical stability issue and how it was fixed. That is, the 252 commits investigated in the paper have 100\% agreement.
 

\subsection{Constructing DeepStability}
The 252 numerical stability commits are related to mostly patches (187 commits) and unit tests (38 commits). There are also new features (8 commits) and speed optimizations (4 commits). {\it others} includes cases such as logging and exceptions related to numerical stability. Among the 252 commits, 137 are related to C/C++, 83 are related to Python, and 22 are related to CUDA. The rest are mostly related to mixed languages, e.g.,  C++ and CUDA.

We performed an in-depth analysis of all 252 commits with the goal to identify instability patterns, root causes, their impact on DL algorithms, and their solutions such as patches and unit tests. We constructed a continuously growing database called {\it DeepStability} that documents our data. It contains 21 columns with the important ones being \textit{Index, Library, Commit hash, Language, Type of commit, Root Cause, Manifestation/End User Impact, IEEE arithmetic exception type, Background, Problem, DL Topics, Patch types, Old Solution, New Solution, and Unit test}. 



{\it DeepStability} is publicly available at \url{https://deepstability.github.io} and can serve as a repository that collects unstable numerical methods and their solutions for machine learning to allow developers to learn how to implement and fix these methods. We plan to continuously contribute to and improve this database in future work. In this paper, we ourselves used this data for further analyses, classifications and summaries, and answered the three research questions shown as follows.






\section{Research Questions and Results}
\subsection{RQ1: Which DL algorithms are susceptible to numerical instability?}

Numerical instabilities are hard to detect once introduced. Localizing which algorithms are susceptible to numerical instabilities can inform developers to be especially careful when implementing these algorithms. In addition, diagnosing DL failures is often challenging, as there can be many factors that lead to ineffective learning, e.g., inadequate dataset, incorrect hyper-parameters or numerical instability. Moreover, it is hard to pinpoint where numerical issues in an algorithm implementation originate from. In Table~\ref{DLTopic_detail}, we provide a list of DL algorithms and numerical methods where we found numerical instability. We hope this list can help developers to narrow down where to inspect DL code to detect and diagnose numerical instabilities. 

As shown in the first column in Table~\ref{DLTopic_detail}, these algorithms and methods belong to a variety of topics, ranging from  well-known DL components of activation functions, loss functions, CNN operations, optimizers and data processing to lower level learning implementations such as tensor math, derivatives, statistical distributions and linear algebra, as well as performance aware learning that involves low precision calculations (i.e.: less than 32 bits) for faster execution such as quantization and other non-standard precision training. 

Under {\it Count} and {\it \% of Total}, we show that tensor math (e.g.: the computation of log, exp, sum and power on tensor), statistical distributions (e.g.: computing log probability, sampling, precision matrix), and data processing (e.g., batch normalization) report the most frequent problems. Please note Table~\ref{DLTopic_detail} shows only numerical instabilities located in DL implementations, which accounted for 88\%. The remaining 12\% of commits were numerical stability in DL implementations but not related to DL algorithms, e.g., overflow when performing timing. 

\begin{table*}[ht]
\caption{DL Algorithms and Methods Susceptible to Numerical Instability}~\label{DLTopic_detail}
\begin{tabular}{llll}
\hline
\textbf{Topic}      & \textbf{DL Algorithms and Numerical Methods}                                                                                                                                                                                                       & \textbf{Count} & \textbf{\% of Total} \\ \hline 
Tensor math                     & \begin{tabular}[c]{@{}l@{}}summation, variance, remainder, mean, standard deviation, sum of squares, \\ log approximation, range, division, power, exponential\end{tabular}                                                                        & 38             & 15\%                 \\ \hline
Statistical distributions       & \begin{tabular}[c]{@{}l@{}}Gaussian, Binomial, Multivariate normal, Laplace, Gumbel, Gamma, \\ Dirichlet, Poisson, precision matrix, sampling, log probability\end{tabular}                                                                        & 26             & 10\%                 \\ \hline
Data processing                 & \begin{tabular}[c]{@{}l@{}}batch normalization, parallel training, tensor shape, \\ tensor allocation, image processing\end{tabular}                                                                                                               & 22             & 9\%                  \\ \hline
Quantization                    & quantization aware training, dequantization                                                                                                                                                                                                        & 17             & 7\%                  \\ \hline
Linear algebra                  & determinant of a matrix, norms, cosine similarity distance                                                                                                                                                                                         & 16             & 6\%                  \\ \hline
Activation functions            & \begin{tabular}[c]{@{}l@{}}leky relu, softmax, logsoftmax, sigmoid, logsigmoid, \\ spatial logsoftmax, softplus, PRelu\end{tabular}                                                                                                                & 13             & 5\%                  \\ \hline
Non-standard precision training & mixed precision, half precision, ultra low precision, precision conversion                                                                                                                                                                         & 11             & 4\%                  \\ \hline
Derivatives                     & gradients                                                                                                                                                                                                                                          & 11             & 4\%                  \\ \hline
Loss functions                  & \begin{tabular}[c]{@{}l@{}}binary cross entropy loss, cross entropy loss, \\ poisson negative log likelihood loss, logistic loss\end{tabular}                                                                                                      & 10             & 4\%                  \\ \hline
CNN operations                  & \begin{tabular}[c]{@{}l@{}}max pooling, LP Pooling,average pooling, \\ convolution transpose, rotated triangle intersection\end{tabular}                                                                                                           & 10             & 4\%                  \\ \hline
Optimizers                      & SGD, Adagrad, centered RMSprop                                                                                                                                                                                                                     & 6              & 2\%                  \\ \hline
Other DL operations             & \begin{tabular}[c]{@{}l@{}}linear interpolation, inverse hyperbolic sine, random number generator, \\ bucket sort, computational graph, csiszar divergence, sparse operations,  \\ word to vec embedding, external libraries (Caffe2)\end{tabular} & 42             & 17\%                 \\ \hline
Total                                            &                                                                                                                                                                                                                                                    & 222   & 88\%       \\ \hline

\end{tabular}
\end{table*}

\subsection{RQ2: What is the cause and impact of numerical instability in DL algorithms?}



Several commits (e.g.: index 55, 61, and 27 in {\it DeepStability}) in this dataset are tracked as high priority bugs. Table \ref{errors} shows that the most common errors in software caused by numerical instability are overflow (47\%) and loss of precision (34\%) followed by underflow (16\%). The rows where multiple errors are listed such as {\it overflow, underflow} indicate that numerical instability can lead to different code errors depending on failure-inducing inputs. There are also commits that amend incorrect comments about stability or optimize the speed of code related to stability, which we list in {\it N/A}.

\begin{table}[h]
\caption {Errors Caused by Numerical Instability} 
\label{errors}
\begin{tabular}{lll}
\textbf{Errors in code}                  & \textbf{Count} & \textbf{\% of Total} \\ \hline
overflow                             & 118            & 46.8\%                 \\ \hline
loss of precision                    & 86             & 34.1\%                 \\ \hline
overflow, underflow                   & 19             & 7.5\%                  \\\hline
underflow                            & 16             & 6.3\%                  \\\hline
overflow, loss of precision           & 3              & 1.2\%                  \\\hline
underflow, loss of precision          & 1              & 0.4\%                  \\\hline
overflow, underflow, loss of precision & 1              & 0.4\%                  \\\hline
invalid input                   & 2              & 0.8\%                  \\\hline
N/A             &6     &2.4\%\\\hline
Total                                & 252            & 100\%                \\ \hline
\end{tabular}
\end{table}

We observed that in a neural network, loss of precision can cause inaccurate updates to its weights and biases and therefore, inferior learning. Overflow and underflow produce values that are equal to inf and 0 respectively, which will cause NaNs in the model parameters. When the model parameters become \textit{NaN}, the model cannot learn and any further code execution is a waste of computational resources and the software engineer's time. \textit{NaN} outputs should be easy to detect, yet we find that DL APIs such as Keras continue executing training even when loss and gradients become \textit{NaN}.


Table \ref{manifestation} shows examples of numerical instability manifestations we discovered. Specifically, it shows inputs to various algorithms that trigger incorrect or inaccurate outputs, which are demonstrated by comparing the actual and expected outputs. As shown in row 1 of Table \ref{manifestation}, an unstable implementation of log determinant of a matrix outputs -inf given an input matrix with 512 rows and 512 columns and small entries equal to 2e-7, while the expected correct output equals -6718.6489. Row 2 of Table \ref{manifestation} gives an example of a unstable remainder calculation, which yields an incorrect result of 128 for 2749682432.0 modulo 36, which is very far from the correct result of 20. Row 3 shows that a numerically unstable implementation of cosine similarity distance may return a value greater than 1.0, which is incorrect because cosine similarity is defined on a range from -1.0 to 1.0. Row 4 shows that an unstable implementation of log probability can yield -inf for binomial distribution initialized with large logits. The reason is that the intermediate calculation that involves multiplication of logits and number of Bernoulli trials overflows and is set to inf. The correct output is 0, because $logit = 90.5229$ corresponds to a probability very close to 1 and log(1) = 0.


\begin{table*}[h]
\caption {Examples of Numerical Instability Failure Inducing Inputs and Manifestation} 
\label{manifestation}
\begin{tabular}{llll}
\hline
\textbf{algorithm}                       & \textbf{failure inducing input}                                                                 & \textbf{output}   & \textbf{expected output} \\ \hline
matrix log determinant                   & 512 by 512 matrix with elements equal to 2e-7                                                   & -inf              & -6718.6489               \\ \hline
remainder                                & 2749682432.0 \% 36                                                                              & 128               & 20                       \\ \hline
cosine similarity & \begin{tabular}[c]{@{}l@{}}u = {[}13.189142, 8.138781, ..., -4.0982385, 5.143065{]}\\ v = {[}13.188879, 8.138888, ..., -4.0983186, 5.1430016{]}\end{tabular} & 1.0000002 & slightly less than 1.0                      \\ \hline
log probability of binomial distribution & logits = 90.5229                                                                                & -inf              & 0                        \\ \hline
\end{tabular}
\end{table*}

\subsection{RQ3: What solutions are used for handling numerical instability in DL algorithms?}
Our goal here is to discover and summarize solutions of numerical instability in DL implementations. These solutions can be directly used by developers to handle similar numerical instabilities and also  serve as a starting point for solving new numerical instabilities in DL.

We identify a list of solution patterns, shown in Table~\ref{solution}. There are four primary categories of solutions for fixing unstable implementations: (1) {\it rewriting math formula}, (2) {\it increasing precision or change variable type}, (3) {\it using a different algorithm} and (4) {\it  limiting input range}. In addition to there four solution types, Table~\ref{solution} also lists {\it mixed precision training}, which allows for speeding up computationally intensive neural network training. The remaining solutions in Table~\ref{solution} pertain to detection such as adding overflow check into algorithm implementations and adding or fixing assertions and unit tests. Interestingly, we observe that some assertions, tests and arithmetic exceptions are ignored as shown in Table~\ref{solution} as {\it relax accuracy test tolerance}, and {\it ignore test/error messages}. In the following subsections, we provide further details on the top four solutions.



\begin{table}[h]
\caption {Numerical Stability Solution Patterns}~\label{solution} 
\begin{tabular}{lll}
\textbf{Solution Type}                              & \textbf{Count} & \textbf{\% of Total} \\ \hline
rewrite math formula                             & 63             & 25.00\%              \\ \hline
increase precision/change variable type & 59             & 23.41\%              \\ \hline
use a different algorithm                        & 43             & 17.06\%              \\ \hline
limit input range                                & 21             & 8.33\%               \\ \hline
relax accuracy test tolerance                    & 14             & 5.56\%               \\ \hline
add overflow check                              & 14             & 5.56\%               \\ \hline
add/fix assertion or unit test                                & 13              & 5.16\%               \\ \hline
ignore unit test/exceptions                            & 12             & 4.76\%               \\ \hline
mixed precision training                         & 6              & 2.38\%               \\ \hline
other                                            & 7              & 2.78\%               \\ \hline
Total                                      & 252            & 100.00\%             \\ \hline
\end{tabular}
\end{table}

\vspace{-15pt}
\subsubsection{Rewriting mathematical formula} 

We distilled a list of templates of unstable math formulas and their solutions, shown in Table \ref{table_formula_patterns}. We find that three approaches are often used to rewrite mathematical formulas for improving stability: (1) {\it using different operations}, (2) {\it re-ordering operations}, or (3) {\it adding a small epsilon}. Row 1 in Table \ref{table_formula_patterns} shows an example, where a mathematical formula can be rewritten to use different operations to improve numerical stability. Square root can suffer from loss of precision for small inputs and multiplying two values that suffer from loss of precision yields a result with even greater precision loss. A better solution is to avoid that and take a square root of $x^2$, which should yield exactly x. Row 2 in Table \ref{table_formula_patterns} shows an example, where a different order of operations can improve numerical stability. Instead of subtracting the sum of max and log(y) from x, we should first subtract max from x and then log(y) to avoid subtraction of two numbers with very different magnitudes that leads to loss of significant digits. Row 3 in Table \ref{table_formula_patterns} is an example, where adding a small epsilon to the input value of log prevents invalid operation log(0), which is undefined and results in an arithmetic exception.
Deriving various mathematically equivalent formulas to find a numerically stable solution can be challenging, which we demonstrate with the example below discovered in this study.


\begin{table*}[h]
\caption{Rewriting Mathematical Formulas to Improve Numerical Stability}
\label{table_formula_patterns}
\begin{tabular}{llll}
\textbf{Vulnerability template} & \textbf{Patch template} & {\textbf{Fix}} &  \textbf{Impact}     \\\hline 
x / (sqrt(x) $*$ sqrt(x)) & x / sqrt(x $*$ x) & use different operations & loss of precision, or incorrect result\\\hline
x - (max + log(y)) & x - max -log(y) & rewrite order of operations & loss of significant digits \\\hline
x - y $*$ log(x) & x - y $*$ log(x+epsilon) & add small epsilon to log & invalid operation if x = 0\\\hline
x * y/ (z*z) &  x * (y/z)/z & use different operations & overflow/underflow if z large/small \\\hline
x + epsilon + $y^{2}$ & x + $y^{2}$ + epsilon & rewrite order of operations & underflow not prevented\\\hline 
(x + y - 1) / t & (x - 1) / y + 1 & rewrite order of operations & overflow if x is large\\\hline
(x + y) / 2 & x + (y - x) / 2 & rewrite formula & overflow\\\hline
(8 $*$ x $*$ y + 31) / 32 $*$ 4 & (x $*$ y + 3) / 4 $*$ 4 & rewrite formula & overflow\\\hline
\end{tabular}
\end{table*}

\noindent{\bf Example 5.1 [Synchronized Batch Normalization]} Synchronized batch normalization (SyncBN) is a type of batch normalization used for paralleled neural network training that utilizes multiple GPUs, where each mini-batch of data is divided across multiple GPUs that calculate the gradients used for updating weights and biases. Batch normalization is a form of data processing that scales inputs to conform to the standard normal distribution that has a mean of 0 and a standard deviation of 1. It enables faster and more stable training of DNNs \cite{Santurkar2018HowDB}. Standard batch normalization only normalizes the data within each GPU, while SyncBN normalizes the inputs within the whole mini-batch as opposed to different mini-batch subsets that reside on different GPUs. SyncBN is defined in Equation \ref{eq_1} ~\cite{Ioffe2015BatchNA}, where $\gamma$ and $\beta$ are learnable parameter vectors of length equal to the input size, and are initialized to uniform distribution on (0, 1) interval and 0 respectively.

\vspace{-5pt}


\begin{equation} \label{eq_1}
\frac{x - E[x]}{\sqrt{Var[x] + \epsilon}} * \gamma + \beta
\end{equation}

In our study, we observed a numerically unstable implementation shown in Equation \ref{equation2}.
This implementation loses precision due to the \textit{floor} operation, and is also slow due to performing unnecessary power and log computations, which are expensive.



\begin{equation} \label{equation2}
2^{floor(log_2(\frac{1 + (x - E[x])-1}{\sqrt{Var[x] + \epsilon}}))}
\end{equation}   

More numerically stable  and faster solution is shown in Equation \ref{equation3}, which is equivalent to Equation \ref{eq_1} for $\gamma = 1$ and $\beta=1-\frac{1}{\sqrt{Var[x] + \epsilon}}$.

\begin{proof}
\small
\begin{align*}
    \frac{x - E[x] + \sqrt{Var[x] + \epsilon} - 1}{\sqrt{Var[x] + \epsilon}}\\
    = \frac{x - E[x]}{\sqrt{Var[x] + \epsilon}} + \frac{\sqrt{Var[x] + \epsilon}}{\sqrt{Var[x] + \epsilon}} - \frac{1}{\sqrt{Var[x] + \epsilon}}\\
    = \frac{x - E[x]}{\sqrt{Var[x] + \epsilon}} + 1 - \frac{1}{\sqrt{Var[x] + \epsilon}}    
\end{align*}
\end{proof} 

\begin{equation} \label{equation3}
\frac{x - E[x] + \sqrt{Var[x] + \epsilon} - 1}{\sqrt{Var[x] + \epsilon}}
\end{equation}    

         
\subsubsection{Increase precision or change variable type}
Risk of overflow and underflow can be mitigated by increasing variable precision or changing its type. We often see that numerical stability is increased by changing a float to a double and an int32 to int64. Changing a variable type from signed int to an unsigned int increases precision and also prevents undefined behavior of signed integer overflow or underflow. In Table~\ref{table_rq2}, we present 9 concrete examples in different scenarios of DL implementations. Under {\it Patch}, we provide a diverse set of type changes that have been utilized to fix numerical stability in DL. The first row in Table~\ref{table_rq2} shows an example of increasing the precision of intermediate calculations from float16 to float32 to prevent NaN and inf gradients in character embedding that is used in NLP. Row 2 is another NLP example, were the precision of a variable that holds the text corpus size is increased from int32 to int64 to prevent overflow of large text files.
Row 3 shows an example, where the accuracy of 2D convolution in quantization aware training is improved by increasing the precision of all variables in the calculation from float to double, i.e. from 32 bits to 64 bits.
\begin{table*}[]
\caption{Changing Variable types or Increasing Precision to Improve Numerical Stability}
\label{table_rq2}
\begin{tabular}{llll}
\hline
\textbf{Algorithm}                                                     & \textbf{Operation}  & \textbf{Problem}                                                                                                             & \textbf{Patch}                                                                                                     \\ \hline
NLP                                                                    & character embedding & \begin{tabular}[c]{@{}l@{}}during FP16 training, character embedding \\ weights receive NAN or INF gradients\end{tabular}    & \begin{tabular}[c]{@{}l@{}}Use a float32 for intermediate results \\ when the input is float16\end{tabular}        \\ \hline
NLP                                                                    & Word2Vec embedding  & text files larger than 2B words overflows                                                                                    & \begin{tabular}[c]{@{}l@{}}increase precision of corpus size \\ from int32 to int64\end{tabular}                   \\ \hline

\begin{tabular}[c]{@{}l@{}}quantization \\ aware training\end{tabular} & 2D covolution       & \begin{tabular}[c]{@{}l@{}}backward pass output in quantization aware \\ training is not accurate enough\end{tabular}        & \begin{tabular}[c]{@{}l@{}}Increase precision of all variables from \\ float to double\end{tabular}                \\ \hline

\begin{tabular}[c]{@{}l@{}}random number \\ generator\end{tabular}     & range               & signed integer overflow of variable range                                                                                    & \begin{tabular}[c]{@{}l@{}}change type of variable range from signed \\ to unsigned int 64 bits\end{tabular}       \\ \hline
summation                                                              & index               & for loop index overflow if input vector is large                                                                             & increase precision from int to int 64                                                                              \\ \hline
flatten layer                                                          & size                & overflow of flattened layer tensor                                                                                           & \begin{tabular}[c]{@{}l@{}}increase precision of variable shape \\ from int32 to int64\end{tabular}                \\ \hline
statistics                                                             & mean                & overflow of sum in mean calculation                                                                                          & Upcast int8, int16, int32 into int64                                                                               \\ \hline
tensor math                                                            & division            & \begin{tabular}[c]{@{}l@{}}division, where the denominator is a low \\ precision scalar has a risk of underflow\end{tabular} & \begin{tabular}[c]{@{}l@{}}Replace the type used for accumulation \\ to the same type as the operands\end{tabular} \\ \hline
optimizer                                                              & parameter size      & \begin{tabular}[c]{@{}l@{}}the size of iterable that holds model parameters\\ has a risk of overflow\end{tabular}            & change int to size\_t \\ \hline                                                                                            

\end{tabular}
\end{table*}


\subsubsection{Use a different algorithm}
Numerical instability can be alleviated by solving a problem with a different algorithm. This patch can involve performing intermediate steps or the entire solution with a different algorithm. We summarize our findings in Table \ref{table_algo_patterns} regarding algorithm choices for a set of common DL operations. The following Example 5.2 below refers to row three of Table~\ref{table_algo_patterns}.

\begin{table*}[h]
\caption{Using Different Algorithms to Improve Numerical Stability}
\label{table_algo_patterns}
\begin{tabular}{lll}
\textbf{DL operation}  & \textbf{Numerically unstable algorithm} & \textbf{Numerically stable algorithm}       \\\hline

matrix inverse  & direct matrix inverse & Cholesky inverse\\\hline   
variance, standard deviation & \begin{tabular}[c]{@{}l@{}}naive algorithm based on variance definition,\\ two-pass algorithm\end{tabular} & Welford's algorithm \\\hline
gradient approximation & asymptotic approximation  & Taylor series expansion, Rice saddle point expansion\\\hline  
summation  & sum in any order & sum from smallest to largest\\\hline  

statistical distributions & parametrize with probabilities & parametrize with logits\\\hline                                                            
statistical distributions  & compute determinant & compute log determinant\\\hline 
loss and sigmoid & compute loss then apply sigmoid & combinine sigmoid and BCE loss into one layer\\\hline  
\end{tabular}
\end{table*}

\noindent{\bf Example 5.2 [Gradient Approximation in Gamma Distribution]} 
The gamma distribution is a continuous distribution that is parametrized with a shape parameter $\alpha$ and rate parameter $\beta$, which determine the shape and range of the distribution respectively. The rate parameter $\beta$ determines the range of the distribution in terms of how stretched or compressed it is. The gradient wrt to $\alpha$ has no analytical form, which is why it needs to be approximated. Depending on the magnitude $\alpha$ and input $x$, different techniques should be used to achieve high precision gradient approximation~\cite{Knowles2015StochasticGV}. For a small $\alpha << 1$, asymptotic approximation yields an accurate gradient, but for a large $\alpha >> 1$ Rice's saddle point expansion~\cite{Lugannani1980SADDLEPA} should be used. For small inputs $x$, Taylor series instead of asymptotic approximation should be used.

\subsubsection{Limit input range}
Numerical instability can be prevented by adding {\it bounds-check}, i.e., imposing restrictions on what maximum and minimum inputs are allowed to perform this computation.

\noindent{\bf Example 5.3 [Uniform Distribution Number Generator]} Uniform distribution has a constant probability and is defined by an interval $[a, b]$, where $a$ is the minimum and $b$ is the maximum value that can occur.  Generating random uniform numbers between -0.9 and 1.0 leads to creating {\it denormalized numbers}~\cite{Schwarz2003HardwareIO}, which are very small numbers that are close to 0, and must be represented with a zero exponent and a mantissa whose leading bit(s) are zero. Denormal number computations are not only slower, but also lead to loss of precision and therefore, a risk of underflow. To avoid that, random uniform numbers should be generated on an interval between 1 and 1.125, which produces a {\it normalized} range of values.

\section{Newly discovered unstable methods and their solutions in DL}
We select several interesting newly discovered numerical instabilities and discuss the details of their problems and solutions along with contributions to existing literature. The rest can be found in DeepStability at \url{https://deepstability.github.io}.

\subsection{Cosine Similarity}
Cosine similarity (Index 4 in DeepStability) is a measure of the angle between two non-zero vectors and therefore, it represents how similar are the directions of the two vectors. Two cosine vectors that have the same direction (regardless of their magnitude) have an angle of 0 degrees between them and therefore have a cosine similarity of 1. If the vectors are pointing in opposite directions, a 180 degrees angle, their cosine similarity is -1. And finally, two orthogonal vectors sharing an angle of 90 degrees have a cosine similarity of 0. Therefore, cosine similarity is defined as follows: 
\begin{equation}
\label{cossim}
    cos\_sim = cos(\theta) 
    = \frac{\vec{u} * \vec{v}} {\|u\| * \|v\|}
    = \frac{\sum_{i=1}^{N} u_{i}*v_{i}} {\sqrt{\sum_{i=1}^{N} u_{i}^{2}} * \sqrt{\sum_{i=1}^{N} v_{i}^{2}}}
\end{equation}

In our study, we identified the numerically stable and unstable versions of cosine similarity (to our best knowledge, there is no prior literature that discusses numerical instability of cosine similarity), shown in Algorithm \ref{cosine_sim_algo}. Lines 4-7 lead with '+' in blue indicate stable code and lines 8--10 with '-' in red show unstable code. The root cause of instability is the inverse square root operation at line 8 and shown in Equation \ref{equation_cos_sim_unstable}. The following mathematical formulas are mathematically identical, but Equation \ref{equation_cos_sim_unstable} is less numerically stable than Equation \ref{equation_cos_sim_stable}. 
\vspace{-2pt}
\begin{equation} \label{equation_cos_sim_unstable}
x * \frac{1}{\sqrt{y*z}}
\end{equation}

\begin{equation} \label{equation_cos_sim_stable}
\frac{x}{\sqrt{y*z}}
\end{equation}

A numerically unstable implementation of cosine similarity distance may return a value greater than 1.0, which is incorrect, because cosine similarity can only range from -1.0 to 1.0. 
Cosine similarity is used, for example, in NLP for measuring similarity between vector representations of text for document classification. Given two documents, a cosine similarity of 1 implies that they are precisely the same and a cosine similarity of 0 means that they are completely different. An unstable implementation of cosine similarity that yields wrong results can therefore lead to incorrect text classification.

\begin{algorithm}
	\caption{Numerically Stable vs Unstable Cosine Similarity Algorithm} 
	\label{cosine_sim_algo}
	\LinesNumbered
	\KwIn{$\vec{u}$, $\vec{v}$, espilon}
	\KwOut{cosine similarity of $\vec{u}$ and $\vec{v}$}
    $x = sum(\vec{u} * \vec{v})$\\
    $y = sum(\vec{u} * \vec{u})$\\
    $z = sum(\vec{v} * \vec{v})$\\
    {\color{blue}+ $n = y * z$}\\
    {\color{blue}+ clamp n to ensure n $>=$ (epsilon * epsilon)}\\
    {\color{blue}+ $n = sqrt(n)$}\\
    {\color{blue}+ $result = x / n$}\\
    {\color{red}- $n = 1/(sqrt(y * z))$}\\
    {\color{red}- clamp n to ensure n $<=$ (1.0/epsilon)}\\
    {\color{red}- $result = x * n$}\\
    \Return {result}
\end{algorithm}

\subsection{Bucketization Algorithm}
Bucketization algorithm (Index 28 in DeepStability) categorizes inputs based on boundaries, e.g.: for boundaries $b = [0, 10, 100]$ and input $x = [[-5, 10000] [150, 10] [5, 100]]$, the output is $[[0, 3] [3, 2] [1, 3]]$. Bucketization algorithm leverages binary search, which can cause numerical instability. Binary search is a search algorithm that finds the position of a target value within a sorted array by iteratively comparing the target value and the middle element of the array to cut down the search space in half each time until the target value is found. The midpoint can be calculated using Equation \ref{unstablemidpoint}, where L is the left index initialized to 0 and R is the right index initialized to N-1, where N is the size of the input array.

\begin{equation}
\label{unstablemidpoint}
    midpoint = (L+R)/2
\end{equation}

If N is very large, adding L and R can result in overflow. A more numerically stable solution is Equation \ref{stablemid}.
 
\begin{equation}
\label{stablemid}
    midpoint = L + ((R-L)/2)
\end{equation}
 
Equations \ref{unstablemidpoint} and \ref{stablemid} are mathematically equivalent:
\begin{proof}
    \begin{align*}
        \frac{L+R}{2} &= L + \frac{R-L}{2} = \frac{2L + R - L}{2} = \frac{L+R}{2} 
    \end{align*}
\end{proof} 
\vspace{-2pt}
Equation \ref{stablemid} mitigates the risk of overflow for large values of R and L, because the result of R-L will not be a larger value than R or L. Adding R and L can result in overflow even if R and L are within representable precision range.  For example, given a sorted array between 1 and $2^{-31}-1$ and a target value of $2^{-31}-1$, the unstable search errors out with a segmentation fault in C++ (an integer overflow on an index can cause a read or write outside of bounds of an array, which triggers a segmentation fault).
The stable search correctly outputs that the target value is located in the $2147483646^{th}$ position in the array.
A numerically unstable implementation of Bucketization algorithm may not be able to output any result and yield an error instead. 

Bucketization can be used in feature engineering for transforming numerical features into categorical ones. For example, suppose that we are creating a neural network that predicts house prices, and one of the features are the GPS coordinates. We can leverage bucketization to create a categorical feature that bins all observations into buckets based on defined boundaries. This can boost model accuracy and allows for reasoning about the relationship between the house location and price. Since the unstable implementation cannot output binned values for certain inputs as discussed above, it can decrease the availability or quality of pre-processed input features.
To our best knowledge, there is no literature that identifies numerical stability vulnerabilities in Bucketization algorithm. Using {\it DeepStability}, we have located a numerical stability issue in a binary search implementation in Tensorflow, and submitted a fix~\footnote{https://github.com/tensorflow/tensorflow/pull/50855} which has been accepted and merged in. This new numerical stability issue was found via the same process that we envision for developers to use to benefit from {\it DeepStability}. We observed a fix in PyTorch that involved rewriting the binary search algorithm to improve its numerical stability. We analyzed the issue and solution, and recorded them in {\it DeepStability} as entry 28. We then checked the implementation of binary search in Tensorflow and found that it is numerically unstable. Using the solution recorded in {\it DeepStability}, we submitted a pull request to Tensorflow with a fix and explanation obtained from {\it DeepStability}.

\subsection{Differentiation of the LU Decomposition}
Differentiation of the LU decomposition (i.e.: backward pass of LU decomposition) computes the gradient of matrix $A$ in the LU decomposition (Index 2 in DeepStability), which can be numerically unstable. LU (lower–upper) decomposition (also called LU factorization) factors a matrix $A$ as the product of a lower triangular matrix $L$ and an upper triangular matrix $U$. The elements in the lower triangular matrix $L$ that lie above the diagonal are zero, while in the upper triangular matrix it is the elements below the diagonal are zero. LU decomposition is an efficient method used for solving a system of linear equations.

Differentiation of the LU decomposition requires division by matrix $L$ and $U$. Algorithm \ref{unstable_LU} is numerically unstable, because it relies on an inverting the $L$ and $U$ matrices to perform that division. We discovered a more numerically stable solution from our data shown in Algorithm \ref{algo2}. It replaces the inverse of matrix L and U with solutions to systems of triangular equations. A system of triangular equations has the form of a triangle, because lower equations always contain variables from the equation above, except for the first variable, e.g.: 5x + 4y = 0, 10y - 3z = 11, z = 3. 

\begin{algorithm}
	\caption{LU Backward Numerically Unstable} 
	\label{unstable_LU}
	\LinesNumbered
	\KwIn{L, U, P, LU gradient, pivots gradient}
	\KwOut{gradient of A}
        Create an identity matrix I with shape same as LU\_gradient\\
        $L\_inverse = (triangular\_solve(I, L))^T$ {// unstable~\cite{Higham2002AccuracyAS}}\\
        $U\_inverse = (triangular\_solve(I, U))^T${// unstable~\cite{Higham2002AccuracyAS}}\\
        $\phi_{L} = lower\_triangular(L^T * LU\_gradient)$\\
        Fill diagonal of $\phi_{L}$ matrix with $0s$\\
        $\phi_{U} = upper\_triangular(LU\_gradient * U^T)$\\
        $grad\_perturbed = L\_inverse * (\phi_{L} + \phi_{U}) * U\_inverse$ \\
        \Return{P * grad\_perturbed}
\end{algorithm}
\begin{algorithm}
    \caption{LU Backward Numerically Stable}
    \label{algo2}
    \LinesNumbered
    \KwIn{L, U, P, LU gradient, pivots gradient}
    \KwOut{gradient of A}
        $\phi_{L} = lower\_triangular(conjugate(L^T) * LU\_gradient)$\\
        Fill diagonal of $\phi_{L}$ matrix with $0s$\\
        $\phi_{U} = upper\_triangular(conjugate(LU\_gradient * U^T))$\\
        $\phi = \phi_{L} + \phi_{U}$\\
        $X = triangular\_solve(\phi, conjugate(L^T))$\\
        $A\_grad = conjugate((triangular\_solve(conjugate(X^T) * P^T, U))^T)$\\
        \Return {A\_grad}
\end{algorithm}

The impact of the unstable solution is inaccurate gradient output of differentiation of the LU decomposition. Matrix decomposition is used in image processing for separating the background and foreground of an image or image denoising \cite{Elad2006ImageDV}. In recommender systems such as Netflix it can be used for collaborative filtering \cite{Koren2009MatrixFT} that analyzes relationships between customers and products to identify new associations. In those systems matrix decomposition can be used to characterize customers and products by vectors of factors, which are used to make recommendations. Therefore, an inaccurate backward pass of matrix decomposition can lead to, for example, wrong recommendations and customer dissatisfaction. Prior literature discusses the numerical instability of a matrix inverse, but does not identify or provide a solution for calculating the gradient for LU decomposition.

\subsection{Higher Order Derivatives}
Higher order derivatives (e.g.: in natural gradient descent) perform more than one order of differentiation and can involve division. Division by a large or small value that is squared (Index 3 in DeepStability) can lead to inaccurate results or even overflow or underflow. In the context of higher order derivatives, the formula in Equation \ref{equation4} is applied multiple times. The issue is that calculation of a $n^{th}$ order derivative raises $y$ to the power of $2^n$, i.e.: $y^2$ becomes $y^{2^{n}}$. If $y$ is large or small $y^{2^{n}}$ can overflow or underflow very quickly.

\vspace{-4pt} 
\begin{equation} \label{equation4}
-grad * \frac{x}{y * y}
\end{equation}
\vspace{-4pt} 
\begin{equation} \label{equation5}
-grad * \frac{\frac{x}{y}}{y}
\end{equation}
Instead of dividing by $y*y$, we can divide by $y$ twice as shown in Equation \ref{equation5}. Mathematically $x / y^2 = x / y / y$, but if y is a large finite precision floating point number, then by performing $y^2$ you may lose precision. Successive divisions achieves the same result while not losing as much precision for large values of y.
The impact of the unstable solution is division by inf or zero for large or small values of $y$ respectively. The result of division by {\tt inf} and zero is a zero and {\tt NaN} respectively, which will propagate in the neural network's gradients, weights, biases, and loss and the network will cease learning.

Higher order derivatives are used in natural gradient descent ~\cite{Pascanu2014RevisitingNG} and also in quantum neural networks \cite{Cerezo2021HigherOD}. \cite{Bornemann2011AccuracyAS} offers a theoretical discussion of numerical stability of higher order derivatives of analytic functions.  We provide a practical example of numerical instability and solution for higher order derivatives computations. We find that successive divisions by large or small values that are squared is numerically unstable.



\section{Threats to Validity}
The primary potential internal threat to validity is the  correctness of the analysis of individual commits regarding numerical stability in DL libraries. To mitigate this threat, an independent analysis followed by a discussion was conducted by two of the authors and only items that were fully agreed upon were included.

A potential external threat to validity that our findings may not be representative of all numerical stability vulnerabilities in real-world applications, because we only studied commits in two open-source DL libraries PyTorch and Tensorflow. However, \textit{DeepStability} is a starting point that can serve as a growing repository to continuously record additional numerical stability vulnerabilities and solutions shared by developers and tool builders.

\section{Related Work}
Due to the interdisciplinary nature of the subject, we found that software engineering, numerical analysis and machine learning fields all have developed relevant work.\\ 
\textbf{Numerical bugs and analysis in software}
\cite{Franco2017ACS} conducted an empirical study of numerical bugs in numerical software
libraries: NumPy, SciPy, LAPACK, GNU Scientific Library, and Elemental. They classified four categories of numerical bugs: (1) Accuracy, (2) Special values, (3) Convergence, (4) Correctness. They reported that the most common numerical bug type is correctness and the most common symptom is wrong result followed by crash and bad performance.  Our work studied numerical stability from an algorithm point of view, and focused more on specific unstable numerical methods and their solutions and less on bug statistics. \cite{Franco2017ACS} mention the importance of in-depth domain knowledge for detailed numerical bug analyses, which our work provided.

There have been a set of work on managing numerical errors in software. \cite{Fu2015AutomatedBE} proposed automated backward error analysis techniques for numerical code. \cite{Bao2013OntheflyDO} developed on-the-fly monitoring technique that can predict if an execution of a floating point program is stable. \cite{Lee2015RAIVERA} introduced RAIVE, a tool for detecting instability via identifying output variations of floating point executions. \cite{Yi2019EfficientAR} presented an automated approach to repair floating point errors in numerical libraries via an empirical study of the GSL - GNU scientific library. Their proposed approach involves three steps: error detection using the condition number, approximation extraction, and repair generation. \cite{Chowdhary2020DebuggingAD} introduced a tool for debugging errors in programs using posit representation, an alternative to floating point representation with diminishing accuracy. We hope the detailed numerical stability knowledge we discovered and presented in this work can help improve the above tools.

\textbf{Numerical analysis} Prior research in numerical analysis focuses on theoretical aspects of numerical stability, but does not specifically target DL.
\cite{Higham2002AccuracyAS} is a very comprehensive reference for the behavior of numerical algorithms in finite precision. It covers algorithmic derivations, perturbation theory, and rounding error analysis, which are relevant to both numerical analysis specialists and computational scientists. \cite{Strakos2005OnNS} studied numerical stability of iterative methods in matrix computations such as Jacobi, Gauss-Seidel, and SOR. And \cite{Bousquet2002StabilityAG} defined notions of stability for learning algorithms and show how to use these notions to derive generalization error bounds based on empirical and leave-one-out error.

\textbf{Neural networks} The importance of numerical stability has been mentioned in machine learning textbooks such as  \cite{Goodfellow2015DeepL} (e.g., this book discussed {\it softmax}),  \cite{DeVore2020NeuralNA} analyze neural networks from a viewpoint of mathematics and numerical computation and provide very comprehensive background information. They argue that neural networks are not very numerically stable and use adversarial examples, inputs with very small carefully crafted perturbations that fool the neural network, as evidence to support that claim. \cite{Cohen2020OptimalSN} studies nonlinear methods of approximation and the effects of requiring numerical stability.


\section{Conclusions and future work}
In this paper, we study numerical stability of algorithms and mathematical methods used in deep learning (DL). By analyzing 252 numerical stability commits obtained from PyTorch and TensorFlow, we discovered numerical instabilities in DL methods and their solutions that previous research has not discussed before. We constructed {\it DeepStability}, the first database that catalogs unstable methods, their analyses and solutions for future reuse. We identified a list of vulnerable DL algorithms ranging from well-known DL components such as activation functions, loss functions, CNN operations, optimizers and data processing to lower level learning implementations such as tensor maths and statistical distribution computations and quantization. We found that numerical instability can lead to overflow or underflow and loss of precision depending the input that triggers the vulnerability. These errors impact DL through incorrect or inaccurate results and learning, which lead to unreliable DL models. We provide example inputs that can trigger numerical instability manifestations and analyze the reasons for numerical instability. Finally, we discover and document solutions for numerical instabilities in DL. These include, but are not limited to: rewriting mathematical formulas, increasing precision or changing variable types, using a different algorithm, and limiting input range. In the future, we plan to continue to grow {\it DeepStability} and also implement a web portal to encourage open-source style contributions to it. 
\section{Significance of Contributions}
Numerical stability is very important for robustness and reliability of deep learning, but it is a very hard problem. We have found numerous reports and fixes of numerical stability vulnerabilities in PyTorch and Tensorflow, two very mature DL libraries, which shows that numerical stability is an important issue in DL. Given the increasing demand for DL systems and their use in practice, we need developers to have sufficient knowledge and awareness to prevent, detect, diagnose and fix numerical instabilities.

Despite its importance, numerical stability in DL has not been sufficiently researched due to its challenging interdisciplinary nature. Software engineering (SE) research studied numerical bugs, but not numerical instability in the context of DL algorithms. Numerical analysis works focus on theoretical analysis of maths formulas, but not on code level implementations, patches, failure inducing inputs and unit tests. We observe that machine learning scientists handle numerical stability on a case-by-case basis, but do not aim to consolidate and analyze vulnerability patterns and solutions for reuse. 

Our work explained and reported numerical instability in DL algorithms, provided patches, unit tests, failure-inducing input, and math templates for reuse. We hope that by connecting the three domains, we can enable interesting future research and benefit the three domains. For example, SE researchers can use our findings to design program analyses and tools targeting numerically unstable computations. 


Similar to security vulnerability websites of NVD~\footnote{https://nvd.nist.gov/} and CVE~\footnote{https://cve.mitre.org/}, we created {\it DeepStability} with the goal of continuously documenting numerical stability issues and solutions for future research and for helping developers and tool builders to prevent, detect, localize and fix numerically unstable algorithm implementations. Using  {\it DeepStability}, we ourselves have found and fixed a stability vulnerability in {\it TensorFlow}, and our patch has been accepted by the {\it TensorFlow} team. We believe that that {\it DeepStability} can benefit a broad audience including (1) DL library developers, (2) machine learning/software engineers who use DL libraries, or implement custom DL algorithms, (3) developers of other numerical computational software that implement the same math formulas and algorithms, (4) tool designers who aim to build automatic tools to detect or fix unstable implementations. Indirectly, building numerically stable DL products can benefit many end users.

\bibliographystyle{ACM-Reference-Format}
\bibliography{references}

\section{Appendix}
\subsection{Softmax proof}
Softmax is a commonly used formula in DL that is known to be numerically unstable. Softmax is defined as: 
\begin{equation}
    softmax(x_i) = \frac{e^{x_i}}{\sum_{j=1}^{n} e^{x_j}}
\end{equation}

This formula is numerically unstable and should be implemented as:
\begin{equation}
    softmax(x_i) = \frac{e^{-max(x) + x_i}}{\sum_{j=1}^{n} e^{-max(x) + x_j}}
\end{equation}

The two formulas are mathematically equivalent, which is shown in the proof below.

\begin{proof}
    Let $c$ be a scalar constant such that $log(c) = -max(x_{1}, ..., x_{n})$ \\
    \begin{align}
        softmax(x_i) &= \frac{e^{x_i}}{\sum_{j=1}^{n} e^{x_j}}\\
        &= \frac{c * e^{x_i}}{c * \sum_{j=1}^{n} e^{x_j}}\\
        &= \frac{c * e^{x_i}}{\sum_{j=1}^{n} c * e^{x_j}}\\
        &= \frac{e^{log(c)} * e^{x_i}}{\sum_{j=1}^{n} e^{log(c)} * e^{x_j}}\tag{By property $c = e^{log(c)}$}\\
        &= \frac{e^{(log(c)) + x_i}}{\sum_{j=1}^{n} e^{(log(c)) + x_j}}\tag{By property $e^{a} *  e^{b} = e^{a+b}$}\\
        &= \frac{e^{-max(x) + x_i}}{\sum_{j=1}^{n} e^{-max(x) + x_j}} &&\qedhere
    \end{align}
\end{proof}

\subsection{LogSoftmax proof}
Logsoftmax is anoter canonical example of a numerically unstable formula that is commonly used in DL. Prior literature shows how to rewrite the formula to obtain a numerically stable solution, but does not provide proof that the two formulas are mathematically equivalent. To our best knowledge, this is the fist comprehensive proof which shows that step by step. 

LogSoftmax performs softmax followed by the logarithm function and therefore, outputs log probabilities. LogSoftmax is defined as: 
\begin{equation}
    logsoftmax(\vec{x})_i = \frac{log(e^{x_i})}{\sum_{j=1}^{n} e^{x_j}}
\end{equation}

This mathematical formula is numerically unstable and should be implemented as: 
\begin{equation}
    logsoftmax(\vec{x})_i = x_i - max(\vec{x}) -log(\sum_{j=1}^{n} e^{x_j - max(\vec{x})})
\end{equation}

These two equations are mathematically equivalent, which can be proved utilizing the identity: 
\begin{equation}
\label{identity}
    log(\sum_{j=1}^{n} e^{x_j}) = max(\vec{x}) + log(\sum_{j=1}^{n} e^{x_j - max(\vec{x})})
\end{equation}

We first prove the correctness of the identity and then the mathematical equivalence of the numerically stable and unstable logsoftmax formulas.

\begin{proof}
\small
    Let $c$ be a scalar constant such that $c = max(x_{1}, ..., x_{n})$
    \begin{flalign*}
        log(\sum_{j=1}^{n} e^{x_j})
        &= max(\vec{x}) + log(\sum_{j=1}^{n} e^{x_j - max(\vec{x})})\\
        &= c + log(\sum_{j=1}^{n} e^{x_j - c}) \\ 
        &= c + log(\sum_{j=1}^{n} e^{x_j} * e^{-c})&&\tag{By property $e^{a-b} = e^{a} * e^{b}$}\\
        &= c + log(e^{-c}) + log(\sum_{j=1}^{n} e^{x_j})&&\tag{By property $log(ab) = log(a) + log(b)$}\\
        &= c + (-c * log(e)) + log(\sum_{j=1}^{n} e^{x_j})&&\tag{By property $log(a^{b} = b * log(a)$}\\
        &= c + (-c(1)) + log(\sum_{j=1}^{n} e^{x_j})&&\tag{By property $log(e) = 1$}\\
        &= c - c + log(\sum_{j=1}^{n} e^{x_j})\\
        &= log(\sum_{j=1}^{n} e^{x_j}) \qedhere
    \end{flalign*}
\end{proof} 

\begin{proof}
\small
    \begin{flalign*}
        logsoftmax(\vec{x})_i &= \frac{log(e^{x_i})}{\sum_{j=1}^{n} (e^{x_j})}\\
        &= log(e^{x_i}) - log(\sum_{j=1}^{n} e^{x_j})&&\tag {By property $log(a/b) = log(a) - log(b)$}\\
        &= x_i * log(e) - log(\sum_{j=1}^{n} e^{x_j})&&\tag {By property $log(a^{b}) = b*log(a)$}\\
        &= x_i * (1) - log(\sum_{j=1}^{n} e^{x_j})&&\tag{By property $log(e)=1$}\\
        &= x_i - (max(\vec{x}) + log(\sum_{j=1}^{n} e^{x_j - max(\vec{x})}))&&\tag{By identity in Equation \ref{identity}}\\
        &= x_i - max(\vec{x}) - log(\sum_{j=1}^{n} e^{x_j - max(\vec{x})}) \qedhere
    \end{flalign*}
\end{proof}

\end{document}